\documentclass{Interspeech}

\interspeechcameraready 

\title{Towards few-shot isolated word reading assessment}

\usepackage{microtype}
\usepackage{cite}
\usepackage{tabularx}
\usepackage{array}

\usepackage[normalem]{ulem}

\author{Reuben}{Smit}
\author{Retief}{Louw}
\author{Herman}{Kamper}

\affiliation{Electrical and Electronic Engineering}{Stellenbosch University}{South Africa}
\email{28583272@sun.ac.za, 23742429@sun.ac.za, kamperh@sun.ac.za}
\keywords{automatic reading assessment, child speech, self-supervised learning, few-shot classification}

\usepackage{comment}

\usepackage[prependcaption,textsize=scriptsize]{todonotes}
\definecolor{mycolor}{HTML}{FF6600}

\newcolumntype{C}{>{\centering\arraybackslash}X}
\begin{document}

\maketitle

\begin{abstract}
    %
    We explore an ASR-free method for isolated word reading assessment in 
    low-resource settings. Our few-shot approach compares input child speech to a small set of adult-provided reference templates. Inputs and templates are encoded using intermediate layers from large self-supervised learned (SSL) models. Using an Afrikaans child speech benchmark, we investigate design options such as discretising SSL features and barycentre averaging of the templates. Idealised experiments show reasonable performance for adults, but a substantial drop for child speech input, even with child templates.   Despite the success of employing SSL representations in low-resource speech tasks, our work highlights the limitations of SSL representations for processing child data when used in a few-shot classification system.
\end{abstract}

\section{Introduction}
One of the earliest goals when learning to read is to correctly produce a familiar word when it is presented in isolation.
However, even for this simple oral reading task, automatic assessment systems are only available in a handful of the world's well-resourced languages~\cite{bolanos2013automatic, bernstein2020reading_assessment, silva2021reading_assessment, bai2021iberspeech, harmsen2025reading, yildiz2025reading_assessment}. 
A major challenge is that child speech is inherently difficult to process.
As children develop, the spectral and temporal properties of their speech can vary greatly due to physiological changes, affecting pitch, rhythm, and intonation~\cite{lee1997analysis, yeung2018difficulties, bhardwaj2022childASR}.

The ideal solution is to use a large dataset of labelled child speech in the target language to train automatic assessment models to detect reading errors.
However, child speech is expensive and difficult to collect
\cite{claus2013survey}, especially in low-resource settings.
This calls for creative approaches to develop reading assessment technologies. 
Are there better options to automatically assess isolated word reading for children in low-resource languages? 

Consider a classroom where an educator would like to assess isolated word reading automatically, but no speech model exists for their language. 
The educator can record a handful of examples of the isolated words they would like to test, using their own voice.
Although extremely sparse, producing this small dataset is possible for most educators. 
This data would then be used as templates in an assessment system that presents a test word to a child on a screen, records the child's spoken response, and compares the response to the educator's examples to predict whether the word was read correctly or not.

We present such a few-shot approach. 

\begin{figure}[!t]
  \centering
  \includegraphics[width=\linewidth]{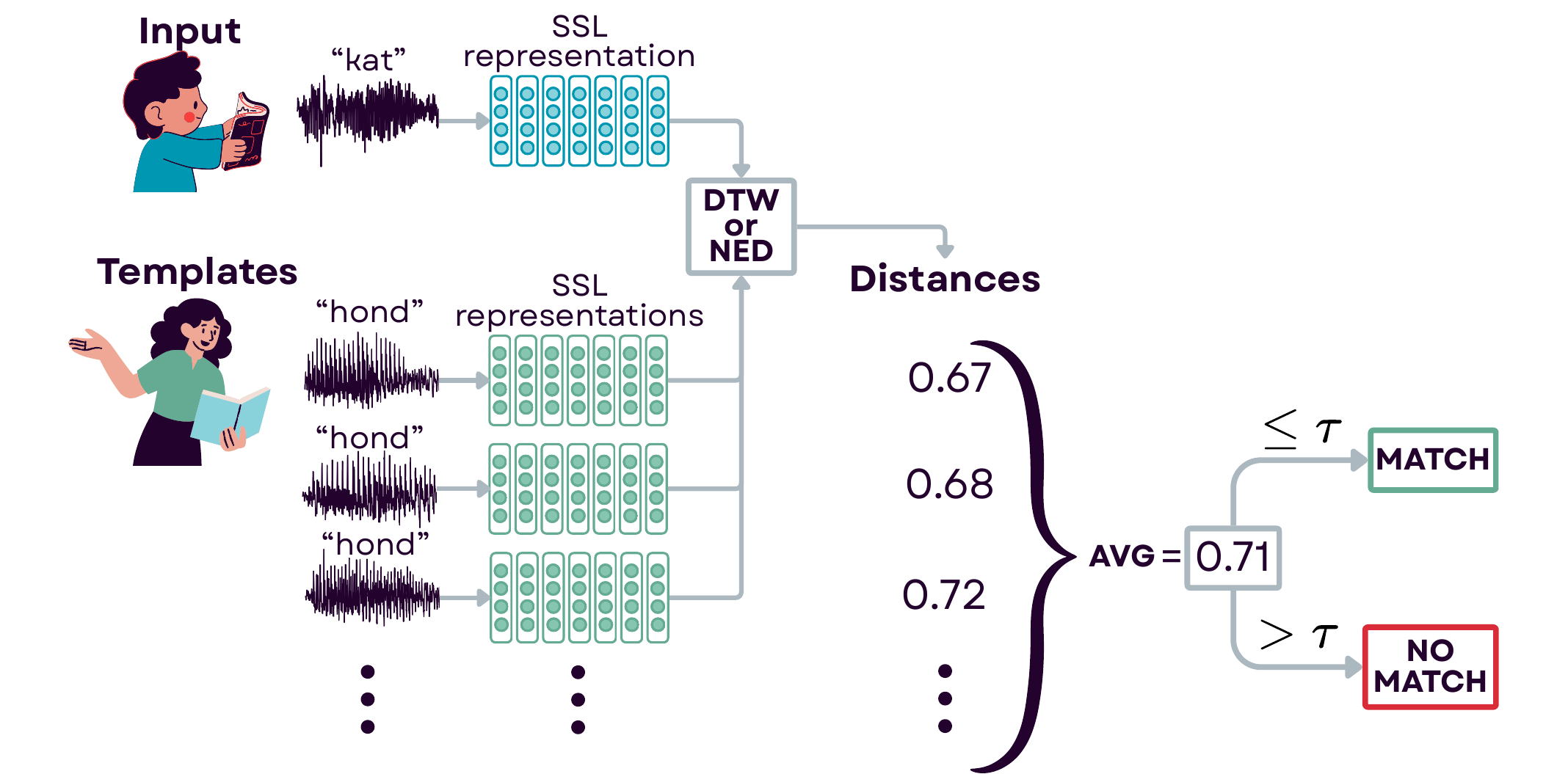}
  \caption{Few-shot isolated word reading assessment. SSL-encoded input from a child is compared to templates from an adult to determine whether the input is correctly read.
  }
  \label{fig:summary}
\end{figure}

Concretely, we simulate this low-resource classroom scenario using
an existing dataset of Afrikaans child speech~\cite{jacobs2025speech} and a new dataset of adult template recordings.
Our few-shot error classification approach is illustrated in Figure \ref{fig:summary}. The input and template speech are encoded using self-supervised learned (SSL) representations from an existing pretrained model.
It has been shown that SSL models can provide robust features, even for unseen languages and domains~\cite{jacobs2021acoustic, conneau2020unsupervised}.
We use these representations to calculate a
distance between each input sample and the template samples.
The input is classified as a correct production of the target word if the distance is below a threshold.
For example, in Figure \ref{fig:summary}, the input
``kat'' has a large average distance from the template representations for ``hond'', and will therefore be classified as an incorrect production if ``hond'' was the target word presented to the child.

We investigate a range of encoding approaches spanning conventional speech features, different SSL models, and output from a mismatched English ASR system.
For SSL representations, we consider both continuous and discretised representations. The latter have been shown to improve speaker invariance in adult speech~\cite{benji2022soft_units}.
Our best approach achieves nearly 12\% higher absolute accuracy than a fine-tuned ASR model.
However, absolute performance is still modest (roughly 65\%) compared to an idealised experiment where adult Afrikaans inputs are used (accuracies higher than 94\%).
Our work therefore shows the promise of using SSL features for an unseen language like Afrikaans in a few-shot reading assessment system, but also calls for improving SSL models for dealing with child speech.

\section{Methodology}

Our overall few-shot assessment approach is illustrated in Figure~\ref{fig:summary}. 
We test different features for encoding the inputs and templates, look at different distance measures based on the types of features used, and then consider ways to aggregate different templates of the same word into a single representative example.

\subsection{Feature encodings}

We consider a range of continuous representations.
A conventional approach is to use mel-frequency cepstral coefficients~(MFCCs)~\cite{abdul2022mel}.
Alternatively, we can use self-supervised learning (SSL) models that are trained to predict masked parts of the input from other unmasked parts.
These models, such as HuBERT~\cite{hsu2021hubert} or WavLM~\cite{chen2022wavlm}, learn to generate frame-wise continuous representations that capture rich phonetic and acoustic structure, making them well-suited for downstream speech tasks. 
In our case, we use mHuBERT~\cite{boito2024mhubert}, a general-purpose multilingual HuBERT model trained on 90k hours of clean multilingual data, including some Afrikaans. 
Although this is mostly adult data, we hope that, since it is exposed to the idiosyncrasies of a variety of languages, it might generalise better to child speech with its acoustic variability.

As an alternative to continuous representations, we consider discrete SSL representations. 
These are sequences of categorical or symbolic units 
derived by clustering SSL features using a K-means codebook. The intuition is that each codebook entry should correspond to a phone-like unit~\cite{vanNiekerk2024spoken}.
We use the index in the codebook to represent frames instead of the multidimensional vectors themselves.
The result is therefore that speech is represented as a sequence of integer codes, such as the ones illustrated later in Figure \ref{fig:discrete_vis}.
In this paper, we specifically use quantised HuBERT-base features, because it was shown to improve speaker invariance for adult speech while preserving phonetic information in \cite{nguyen2022discrete}.

\subsection{Distance measures}

Now that we have representations of our audio, we can determine how similar an input is to the templates. We use one of two dynamic programming algorithms, depending on whether the features are continuous or discrete. For continuous representations (MFCCs and mHuBERT), we use dynamic time warping~(DTW) with cosine distance as the frame-wise metric.
For discrete code sequences (HuBERT discrete), we use edit distance: the sum of the insertions, deletions, and substitutions.
We specifically use normalised
edit distance (NED), which divides the raw edit distance by the length of the longer sequence, resulting in a score between 0 and 1.

\subsection{Barycentres for aggregating templates}

Up to this point, we compared the input to all the templates of a particular word class.
An alternative is to summarise all the templates into a single representative example. 
We use barycentre averaging, a technique where a selected sequence is refined to minimise its distance to all the other sequences. 
There are continuous and discrete variants of this approach.

For continuous features, we use DTW barycentre averaging~(DBA)~\cite{petitjean2011DBA}. 
An initial prototype sequence is first selected by choosing an arbitrary template of median length. 
This avoids outliers in the class templates that may contain noise or incomplete speech. 
Next, the prototype is compared to every other template using DTW, 
mapping individual template features to the frames in the prototype while preserving the sequence order.
This results in a list of vectors corresponding to each frame of the prototype sequence.
The averages of all features mapped to each prototype frame are calculated, giving new prototype frames.
This process is then repeated for a number of iterations, with templates aligned to the updated prototype in each cycle.

For discrete representations, we use the edit distance 
barycentre (EDB) algorithm~\cite{kohonen1985median-strings}. 
EDB also starts by choosing
a prototype sequence of median length. 
After that, we compute the sum of the NEDs
between the prototype and every other sequence in the class, i.e., the total NED. 
Next, a list of neighbours of the prototype sequences is generated by iterating through each index and performing a deletion, insertion, and substitutions of each code in the alphabet of symbolic units. 
We then calculate the total NED between each neighbour and all the sequences in the class -- a costly operation. 
If a neighbour gives a better total NED than the current prototype, it is selected as the new prototype sequence.
This process is repeated until there is no improvement in total NED. 
As an example, see the barycentres of the adult sequences for ``boom'' and ``hond'' in Figure \ref{fig:discrete_vis}.

\section{Experimental setup}
\subsection{Data}

We construct a few-shot evaluation benchmark using an existing dataset of Afrikaans child speech~\cite{jacobs2025speech}.
This dataset contains oral stories from four- and five-year-old children from low-income communities in the Western Cape region of South Africa.
Although Afrikaans is a Germanic language with more than 7 million native speakers, it is still considered a low-resource language because of limited data availability.
We manually extract individual word samples for a selection of word classes.
We divide the samples into training, development, and test subsets, ensuring each subset contains unique speakers.

We first consider evaluation inputs.
16 word classes are treated as target reading words.
For each of these classes, we form a subset of between 6 and 15 positive samples with the same number of negative samples. 
Additionally, we manually select
impostors (i.e., curated negative samples) for three out of these 16 classes.
These imposters are similar but not exact matches to the positive classes and should be classified as incorrect productions.
We select ``muis'' as
impostor for ``huis'', ``seer'' as
impostor for ``nee'', and ``worsies'' as
impostor for ``wors''.
The number of impostors included is equal to the number of positive examples in these classes. 
In total, the development and test evaluation items span about 4 minutes of speech each.

Next, we look at the data that we use for the templates.
We collect 240 adult templates (15 per class) from a single male speaker, spanning roughly 2 minutes.
For analyses under idealised conditions, we also extract a similar set of child templates from the training set of the child speech.
Conversely, we extract adult evaluation inputs for the 13 evaluation word classes and 3 imposter classes; these are obtained from different male and female speakers than the templates.

\subsection{Baseline and topline approaches}
\begin{table}[t]
\renewcommand{\arraystretch}{1.05}
\caption{Examples of English Whisper ASR transcriptions for Afrikaans child and adult speech.}
\label{tbl:english-whisper-examples}
\begin{tabularx}{\linewidth}{@{}lCC@{}}
\toprule
& {``boom''} & {``hond''}\\
\midrule
{Child}   & boom      & and\\
        & one       & and\\
        & hmm       & I am going to go to the\\
        & boom      & honk\\
\vspace{-0.1in}\\
\midrule
{Adult}  & poem  &   boom\\
        & well um   & and\\
        & poem      & and\\
        & poem      & and\\
\bottomrule
\end{tabularx}
\end{table}
One naive baseline is to transcribe both the inputs and templates with a pretrained ASR system.
We assume that we do not have access to an ASR model in the target language, but we do have access to an English ASR model. 
Since this system's language is a mismatch, the output will not be Afrikaans words. 
However, if the system produces consistent ``English''
transcriptions, this could still be useful for comparing an input to templates. 
For this ASR approach, we use English Whisper-base~\cite{radford2023whisper} with NED to compare transcriptions. 
Example output is shown in Table~\ref{tbl:english-whisper-examples}. 

We also consider two more conventional baselines using the templates as training data.
The first is a one-vs-rest softmax regression classifier with single-vector averaged mHuBERT inputs~\cite{pedregosa2011scikit, boito2024mhubert}, trained on the 240 template samples.
The second is a fine-tuned ASR model: we fine-tune Whisper-large-v3-turbo~\cite{polok2025target} on the templates. Whisper~\cite{radford2023whisper} itself has seen adult Afrikaans as part of its pretraining phase, which, together with its high performance in multilingual settings~\cite{torgbi2025adapting}, makes this a strong alternative approach.
Finally, we also test a topline ASR model that is fine-tuned on 5 hours of Afrikaans child speech~\cite{jacobs2025speech}. Such a system will not be available in the majority of low-resource settings, and we therefore use it only to contextualise the results.

\subsection{Evaluation metrics}

Ours is a binary classification task, where a true positive corresponds to correctly predicting that a word was read correctly.
We use a range of metrics for evaluation. In addition to precision, recall, and F1, we also report balanced accuracy:
\begin{align*}
\text{Balanced Accuracy} &= \frac{1}{2}\left(\frac{\text{TP}}{\text{TP}+\text{FN}} + \frac{\text{TN}}{\text{TN}+\text{FP}}\right)
\end{align*}
In contrast to the other metrics, balanced accuracy explicitly accounts for true negatives.
To obtain the classification threshold $\tau$ used for the test set, we evaluate 
different values of $\tau$ on the development set, choosing a $\tau$ resulting in the highest balanced accuracy per representation type.
A single threshold is used for all word classes under that representation. 
Balanced accuracy is calculated per word class and then averaged over the classes.

We also use the receiver operating characteristic (ROC) curve and report the area under this curve (AUC).
ROC plots the tradeoff between the true positive rate,
$\text{TP}/(\text{TN} + \text{FP})$, and the false positive rate, $\text{FP}/(\text{FP} + \text{TN})$, as $\tau$ 
is varied. 
A perfect classifier would follow the top edge of the curve, starting from the far left and continuing to the right~\cite{hosmer2013applied}.
This would give an AUC of 1.0, while an AUC of 0.5 is equivalent to a random~guess.

\section{Experiments and results}

To determine whether SSL representations are a viable option for automatic isolated word reading assessment in a few-shot setting, we examine model performance in three experiments.
The first two are idealised scenarios: we test adult inputs using adult templates, and child inputs with child templates (which would normally not be available).
We then turn to our main experiment, where child inputs are classified using adult templates.

\subsection{Idealised developmental experiments}
\begin{table}[!b]
\renewcommand{\arraystretch}{1.1}
\caption{Isolated word reading assessment results (\%) on the development set.}
\label{tbl:dev-metrics}
\begin{tabularx}{\linewidth}{@{}lCCCCC@{}}
\toprule
Model & Rec. & Prec. & F1 & AUC & Acc. \\
\midrule
\underline{\textit{Adult-adult:}}\\
MFCCs               & 75.6 & 69.7 & 72.6 & 80.9 & 70.6 \\
mHuBERT              & 94.4 & \textbf{96.4} & 95.4 & \textbf{98.6} & 94.5 \\
mHuBERT DBA         & \textbf{95.4} & 96.2 & \textbf{95.8} & 98.4 & \textbf{95.0} \\
HuBERT discrete      & 90.8 & 80.8 & 85.6 & 92.3 & 82.5 \\
HuBERT discrete EDB  & 83.1 & 81.7 & 82.4 & 90.3 & 81.0 \\
English Whisper ASR  & 85.8 & 86.6 & 86.2 & 92.5 & 85.8 \\
\vspace{-0.1in}\\

\underline{\textit{Child-child:}}\\
MFCCs                & 71.7 & 69.7 & 70.7 & 72.8 & 68.8 \\
mHuBERT              & \textbf{87.6} & 72.9 & \textbf{79.8} & \textbf{85.2} & \textbf{76.3} \\
mHuBERT DBA          & 85.3 & 73.6 & 79.0 & 84.7 & 76.2 \\
HuBERT discrete      & 57.1 & \textbf{80.0} & 62.3 & 78.7 & 67.9 \\
HuBERT discrete EDB  & 57.6 & 72.4 & 62.3 & 73.3 & 62.3 \\
English Whisper ASR  & 83.2 & 59.0 & 69.1 & 72.9 & 61.8 \\
\vspace{-0.1in}\\

\underline{\textit{Child-adult:}}\\
MFCCs                & 97.2 & 52.6 & 68.2 & 63.7 & 54.4 \\
mHuBERT              & \textbf{91.8} & 54.2 & 69.1 & 70.6 & 58.9 \\
mHuBERT DBA          & 90.1 & 56.5 & \textbf{69.4} & \textbf{70.9} & 60.0 \\
HuBERT discrete      & 71.7 & 62.7 & 66.9 & 67.7 & \textbf{64.4} \\
HuBERT discrete EDB  & 54.8 & \textbf{66.5} & 59.5 & 66.4 & 64.0 \\
English Whisper ASR  & 80.8 & 58.7 & 67.9 & 70.7 & 61.3 \\
\bottomrule
\end{tabularx}
\end{table}

The results on the development
set are shown in Table \ref{tbl:dev-metrics}.
The adult-adult results serve as an upper bound due to lower acoustic variability.
All feature types perform well, with mHuBERT and mHuBERT with DBA achieving the best scores: F1 and accuracy at around  
95\%, and AUC near 99\%.

This indicates that SSL representations are
effective in the adult domain, even in a low-resource language like Afrikaans. 

The child-child results are consistently lower, reflecting known difficulties in modelling child speech.
mHuBERT again outperforms other representations
with a balanced accuracy of 76.3\%.
Discrete units yield high precision (80.0\%) but low recall (57.1\%), suggesting a conservative decision boundary that fails to identify many true negatives leading to an accuracy of 67.9\%.

In our main child-adult development experiments, performance continues to degrade.
mHuBERT and mHuBERT DBA achieve reasonable F1 scores of 69.1\% and 69.4\%, respectively, whereas mHuBERT DBA achieves the highest AUC, but accuracy drops considerably. 
In contrast to the previous results, we see benefits in using discrete representations, with HuBERT discrete achieving the highest accuracy of 64.4\% (but with low recall and AUC).
These results confirm the difficulty of bridging acoustic and prosodic differences between children and adults, even with state-of-the-art SSL features.

\begin{table}[!t]
\renewcommand{\arraystretch}{1.1}
\caption{Final test set results (\%) using child inputs. Our mHuBERT methods use adult templates.}
\label{tbl:test-metrics}
\begin{tabularx}{\linewidth}{@{}lCCCCC@{}}
\toprule
Model & Rec. & Prec. & F1 &  AUC & Acc.\\
\midrule
mHuBERT                 & 86.5 & 52.6 & 65.4 & 66.0 &\textbf{65.3} \\
mHuBERT DBA             & 58.4 & 54.1 & 66.2 & 65.5 & 56.1 \\
Multiclass regression   & \textbf{100}  & 50.0 & \textbf{66.7} & 20.3 & 50.0 \\
Fine-tuned Whisper ASR& 64.2 & \textbf{69.2} & 66.6 & \textbf{87.9} & 53.4 \\ 
\addlinespace
Topline Whisper ASR (5 hr)    & 72.3 & 68.7 & 70.4 & 86.9 & 53.7 \\
\bottomrule
\end{tabularx}
\end{table}

\subsection{Main experiment and comparison to other systems}

We choose mHuBERT and mHuBERT DBA for comparison to more conventional automatic assessment methods on held-out test data, since they give a reasonable tradeoff between AUC and balanced accuracy in the development experiments in Table~\ref{tbl:dev-metrics}.

The final test results are given in Table \ref{tbl:test-metrics}. 
Among the self-supervised methods, mHuBERT performs competitively with an F1 score of 65.4\% and a high recall of 86.5\%. 
It also achieves the highest balanced accuracy (65.3\%).
The addition of dynamic barycenter averaging (DBA) to mHuBERT does not yield consistent improvements. 

The more standard supervised models that we compare to show varied effectiveness. 
The multiclass regression model predicts that all words are read correctly -- a meaningless prediction -- giving perfect recall and random precision.
This shows the problem with relying solely on F1: artificially inflated recall results in an inflated F1 score.
The Whisper ASR model fine-tuned on roughly 2 minutes of templates achieves the highest AUC (87.9\%). 
But, surprisingly, its balanced accuracy is close to random (53.4\%). The same is true for the topline ASR system trained on 5 hours of child speech; although this system achieves the highest F1 (70.4\%), its balanced accuracy is still low.

Further testing revealed that this discrepancy is because balanced accuracy uses a single classification threshold for all word classes, while AUC calculates the average area as the threshold is varied per word class.
To confirm this, when we tune unique thresholds for each class with the topline system, it achieves a balanced accuracy of 89.9\%, which is in line with its AUC performance.
However, tuning a threshold for each class is not viable in a real-world scenario where it is unlikely to have an existing child development set to tune thresholds for the word classes of interest.
This is why balanced accuracy is a better indicator of a model's performance in our setting.

It is worth emphasising again that, even if we tune thresholds per word class on the topline ASR system, our best child-adult results are still not as good as the best adult-adult results in Table~\ref{tbl:dev-metrics}.  

\subsection{Further analysis}

Our results demonstrate a performance degradation across settings, progressing from adult-adult to child-child, and ultimately to child-adult -- our main focus.
Despite employing techniques such as barycentre averaging to improve robustness, 
performance still suffers under mismatched 
conditions. 
Why?

\begin{figure}[t]
\includegraphics[width=\linewidth]{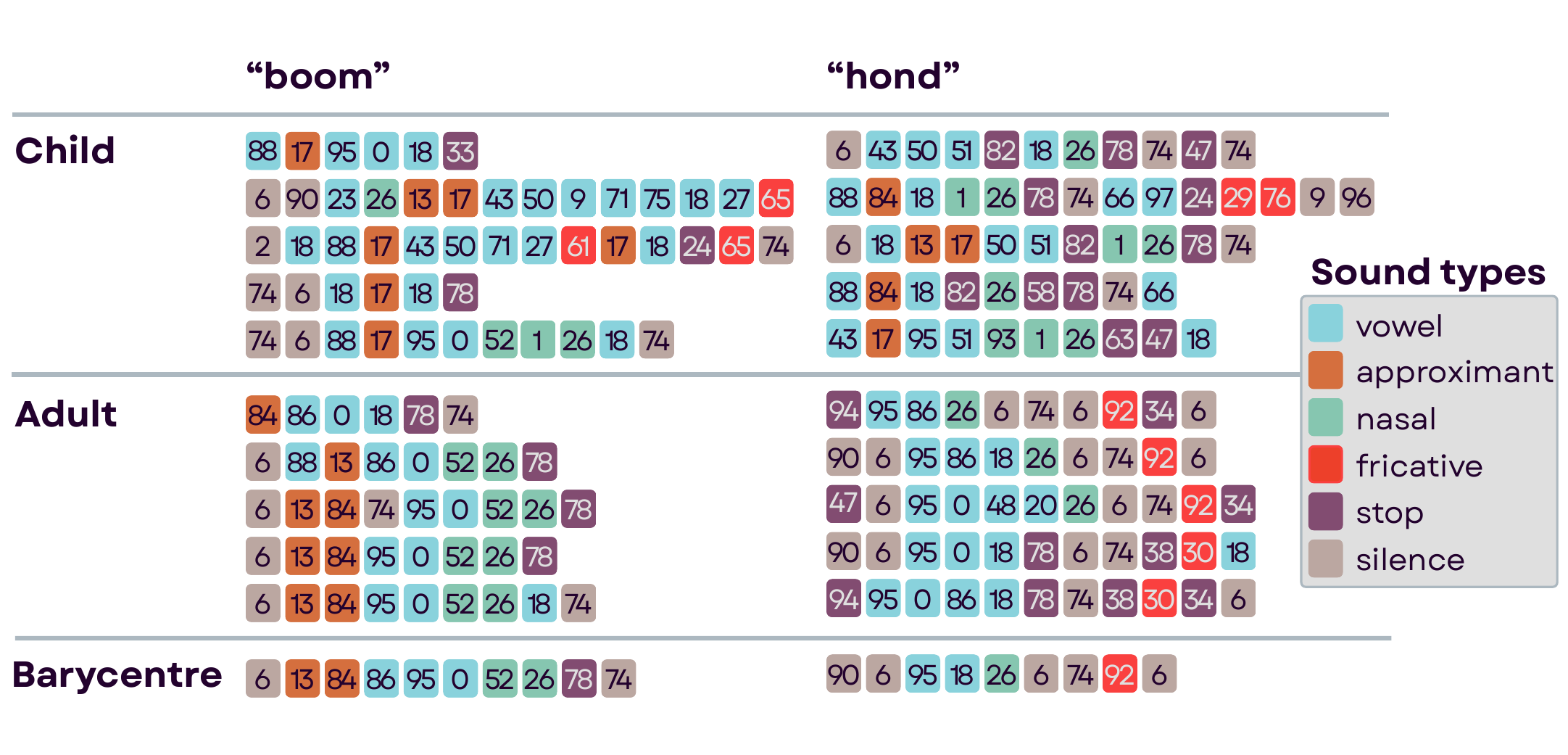}
  \caption{Discrete codes for ``boom'' and ``hond'' for child and adult samples coloured by sound type. 
  }
  \label{fig:discrete_vis}
\end{figure}

\begin{table}[!t]
\renewcommand{\arraystretch}{1.1}
\caption{ROC AUC (\%) for selected classes with mHuBERT features and DTW. Words with a * have specific impostors.}
\label{tbl:ROC-values}
\begin{tabularx}{\linewidth}{@{}lCCCCC@{}}
\toprule
Word & Adult-adult & Child-child & Child-adult\\
\midrule
sien & 100.0 & 100.0 & 91.8\\
water& 100.0 & 93.5 & 82.9\\
wors* & 99.0 & 93.5 & 69.2\\
nee*  & 100.0& 55.6 & 44.4\\
huis* & 83.2 & 51.2 & 41.3\\
\addlinespace
Total & 98.6 & 85.2 & 70.6 \\
\bottomrule
\end{tabularx}
\end{table}

The English Whisper ASR transcriptions in Table~\ref{tbl:english-whisper-examples} show that adult speech yields more consistent outputs than child speech, consistent with prior findings \cite{li2002analysis, wilpon1996study}. 
A similar pattern is evident in the discrete phone unit sequences in Figure~\ref{fig:discrete_vis}, where SSL representations of adult speech are noticeably more stable than those of child speech. 
This consistency holds even across adult speakers with varied characteristics, including a female speaker and one with rhotacism. 
These observations suggest that SSL models are robust to moderate variations in adult speech, such as differences in pitch and articulation, but struggle to handle the greater variability of child speech.
 
Additional insights come from class-level performance patterns shown in Table~\ref{tbl:ROC-values}.
Reading errors for some words, like ``sien'' and ``water'' can be classified fairly robustly. But the classes with specific impostor words such as ``nee'', ``huis'', and ``wors'' show poor AUC in both the child-child
and the child-adult 
settings, 
indicating the model’s difficulty in distinguishing phonetically similar words.
This challenge is less pronounced in the adult-adult setting, where AUC scores are significantly higher.
This reinforces the idea that the core limitation lies in the variability of child speech, rather than the model’s ability to distinguish similar-sounding words in general.

Further work is required to know exactly what properties in the child speech result in more variable encodings.
Apart from the mentioned physiological differences between children, qualitative analyses revealed that our child data are also more variable in the types of background noise and channel conditions.
Again, noise robustness should be considered in SSL representations.

\section{Conclusion}

This paper presented an initial investigation into a few-shot isolated word reading assessment approach for low-resource child speech using SSL representations. 
We introduced a simple, template-based classification method and tested various representations, including continuous and discrete self-supervised speech 
features, barycentre-averaged templates, and a fine-tuned ASR system.
Our findings indicate that while self-supervised models like mHuBERT~\cite{boito2024mhubert}  offer strong performance on adult speech, they are not robust to the variability in child speech, especially when templates are drawn from adults.
Techniques like dynamic time warping barycentre averaging show promise, but require further refinement.
In our follow-up work, we will look into using approaches like voice conversion to align the adult templates more closely to the child inputs~\cite{jacobs2025speech}. 
More broadly, our results suggest a clear path forward: the development of self-supervised models specifically trained or adapted for child speech, ideally across diverse languages and developmental stages. 
This study serves as both a proof of concept and a call to action for the speech community: robust, inclusive models for child speech are necessary to realise equitable educational technologies.

\newpage
\section{Acknowledgements:}
We want to thank Danel Adendorff and Kyle Janse van Rensburg for their assistance with data collection.
This work was supported by a grant from the Het Jan Marais Fonds (HJMF).

\bibliographystyle{IEEEtran}
\bibliography{mybib}

\end{document}